# Title: A Constraint Driven Solution Model for Discrete Domains with a Case Study of Exam Timetabling Problems

Authors: Anuraganand Sharma[1] and Dharmendra Sharma

Corresponding Author: Anuraganand Sharma

Affiliation of both Authors: Information Sciences and Engineering, University of Canberra, ACT, Australia

Emails: sharma_au@usp.ac.fj and Dharmendra.Sharma@canberra.edu.au

Corresponding Author's Phone: +679 3232618 and +679 9143660

**Abstract.** Many science and engineering applications require finding solutions to planning and optimization problems by satisfying a set of constraints. These constraint problems (CPs) are typically NP-complete and can be formalized as constraint satisfaction problems (CSPs) or constraint optimization problems (COPs). Evolutionary algorithms (EAs) are good solvers for optimization problems ubiquitous in various problem domains, however traditional operators for EAs are 'blind' to constraints or generally use problem dependent objective functions; as they do not exploit information from the constraints in search for solutions. A variation of EA, *Intelligent constraint handling evolutionary algorithm* (ICHEA), has been demonstrated to be a versatile constraints-guided EA for continuous constrained problems in our earlier works in (Sharma and Sharma, 2012) where it extracts information from constraints and exploits it in the evolutionary search to make the search more efficient. In this paper ICHEA has been demonstrated to solve benchmark exam timetabling problems, a classic COP. The presented approach demonstrates competitive results with other state-of-the-art approaches in EAs in terms of quality of solutions. ICHEA first uses its inter-marriage crossover operator to satisfy all the given constraints incrementally and then uses combination of traditional and enhanced operators to optimize the solution. Generally CPs solved by EAs are problem dependent penalty based fitness functions. We also proposed a generic preference based solution model that does not require a problem dependent fitness function, however currently it only works for mutually exclusive constraints.

---

[1] Present Address: The University of the South Pacific, School of Computing, Information and Mathematical Sciences

# 1    Introduction

Many engineering problems ranging from resource allocation and scheduling to fault diagnosis and design involve constraint satisfaction as an essential component that require finding solutions to satisfy a set of constraints over real numbers or discrete representation of constraints [23, 24, 58]. There are many classical algorithms that solve CSPs like branch and bound, backtrack algorithm, iterative forward search algorithm, local search but heuristic methods such as evolutionary algorithms (EAs) have mixed success and for many difficult problems these are the only available choice [4, 23, 48]. EAs however suffer from some of its inherent problems to solve CSPs as it does not make use of knowledge from constraints and blindly search in the vast solution space using its heuristic search mechanism. Constraints can reduce the search space and direct the evolutionary search towards feasible regions. Constraint problems (CPs) are divided into two classes: Constrained Optimizing Problems (COPs) and constraint satisfaction problems (CSPs). The difference between these classes is that in the first an optimal solution that satisfies all constraints should be found, while in the second class any solution as long as all the constraints are satisfied is acceptable [30]. Depending on the environment of a problem constraints can be *static* or *dynamic*. Static constraints do not change over time and total number of constraints are also fixed and known *a priori*. Dynamic constraints can change over time which makes the problem more complex like ship scheduling, vehicle routing, dynamic obstacle avoidance, the adaptive farming strategies and aerodynamic/structural wing design problems [24, 49, 58].

Characteristically, the CPs solved by EAs are penalty based fitness functions. A penalty function updates the fitness of chromosomes in EA. A penalty term is used in general for reward and punishment for satisfying and/or violating the constraints [20]. Use of penalty functions has been commonly reported in the literature for use in constrained optimization; however, it has its own advantages and disadvantages that have led to the development of different strategies. Its main advantage is its simplicity and compatibility for EA's objective functions. The major shortfall of the penalty function is that most of them are problem dependent that requires a careful fine-tuning of parameter to obtain competitive results [41]. The penalty factors, which determine the severity of the punishment, must be set by the user and their values are problem dependent [46]. Some other constraint handling approaches include expensive *repair* algorithms that promote the local search to transform infeasible solutions to feasible solutions because the feasible parents not necessarily produce feasible progenies if the search space is non-convex [19]. Some infeasible solution may be very promising like a close neighbor to the optimal solution. To avoid losing such important information repair functions are defined that is also a popular choice for EAs [43]. However its major weakness lies in its problem dependency and in many cases repairing an infeasible solution is itself a complex problem [20]. Some successful repair algorithms are repairGA [49] and GENOCOP III [46]. In Multi-Objective Optimization (MOO), multiple constraints are transformed into multiple objectives. Pareto-based selection approaches are currently the most popular multi-objective evolutionary algorithm (MOEA) solution technique. These solutions are known as *pareto-optimal* solutions or non-dominated solutions [67]. There are many established algorithms like MOGA [32], VEGA [57], NSGA and NSGAII [25] that efficiently solve the constraint problems that can be transformed into multi objective optimization problems. Another approach is hyper-heuristics which are relatively new approach proposed by [9]. These are general systems that are able to handle a wide range of problem domains with respect to meta-heuristic technology which tends to be customized to a particular problem or a narrow class of problems [9]. A hyper-heuristic is an automated methodology for selecting or generating heuristics to solve hard computational search problems [12]. High level heuristic determines how to apply the low level heuristic using problem dependent local properties [5]. Hyper-heuristic with higher level as Genetic algorithm (GA) and lower level as variable neighborhood search (VNS) has been very successful for timetabling problem which is a discrete COP [11]. Some other approaches proposed by Paredis includes co-evolution strategies which unlike the evaluation function of penalty or repair functions, handles constraints and objective separately [52]. It utilizes predator-prey model to keep two populations – one population represents solutions that satisfies many constraints while other population represents those individuals whose constraint(s) is violated by lots of individuals in the first population. Here fitness calculation is expensive as it requires historical record to compute the fitness. Another strategy is based on special representations and operators using decoders technique that maps genotypes to phenotypes [35]. Each decoder links feasible solution and a decoded solution. The idea of this method is to transform a constrained-optimization problem into an unconstrained one by adding (or subtracting) a certain value to/from the objective function based on the amount of constraint violation present in a certain solution. This strategy requires extra computational effort to find the intersection of a line with the boundary of the feasible region.

All the above approaches do not maximally utilize information from constraints. Information discovery from constraints can guide the evolutionary search to improve the performance of EAs as the search operators are 'blind' to constraints [57]. One way is to use error function or distance function from feasible regions. However it is simply one form of a penalty function that is based on the distance of a solution from the feasible region [20, 47]. The advantage of this evaluation function is that it is generic and promotes feasible solution over infeasible one. This approach has been mainly used for continuous COPs. There are some other information discovery techniques as well. Ricardo and Carlos in [55] proposed cultured differential evolution (CDE) that uses differential evolution (DE) as the population space and belief space as the information repository to store experiences of individuals for other individuals to learn. Amirjanov in [2] proposed changing domain range based genetic algorithm (CRGA) that adaptively shifts and shrinks the size of search space of the feasible region by employing feasible and infeasible solution in the population to reach the global optimum. Mezura-Montes et. al. in [41] proposed simple multi-membered evolution strategy (SMES) that uses a simple diversity mechanism by allowing infeasible solutions to remain in the population. A simple feasibility-based comparison mechanism is used to guide the process toward the feasible region of the search space. The idea is to allow the individual with the lowest amount of constraint violation and the best value of the objective function to be selected for the next population. ICHEA uses a constraint guided operator - intermarriage crossover for continuous domain to solve static and dynamic CSPs and COPs in [58, 59, 57, 60]. The intermarriage crossover operator works for non-convex search space as well. ICHEA incrementally adds constraints or set of constraints into the search space which becomes easier to solve than otherwise. ICHEA also extends the feasible region to a degree in a search space to further facilitate locating neighborhoods of feasible region created by a constraint which is then incrementally shrinks towards the actual feasible region.

CSP is defined by an input vector $\vec{x} = \{x_1, x_2, \ldots x_n\}$ of size $n$ in a finite search space $S$ where each variable $x_i$ has a finite domain $D_i$. A set of $m$ constraints $\{c_1, c_2, \ldots c_m\}$ are defined in the form of functions:

$$c_i(x_1, x_2, \ldots x_n) = \begin{cases} 1, if\ satisfied \\ 0, if\ violated \end{cases} \quad (1)$$

Constraint satisfaction sets or feasible regions $S' = \{S_1, S_2, \ldots S_m\}$ created by constraints can also be defined where:

$$S_i = \{\vec{x} \in S \mid c_i(\vec{x}) = 1, i \in \{1, \ldots m\}\} \quad (2)$$

The fitness function of a CSP can be given as violation count:

$$f(\vec{x}) = \sum_{i=1}^{m} c_i(\vec{x}) \tag{3}$$

To incorporate the weighted penalty function Eq. (3) can be redefined as:

$$f(\vec{x}) = \sum_{i=1}^{m} w_i c_i(\vec{x}) \tag{4}$$

Many real world CSPs have constraints defined as qualitative data. The qualitative data are generally represented in discrete form. These CSPs have discrete search space made up of individual points. For example a particular course is scheduled on Monday at a particular room for a timetabling problem. The search space consists of all possibilities of feasible solutions as discrete points with nothing in between them. Searching for solution requires "hopping" from one point to the other unlike continuous search space where the locality of a point, its neighbors and boundaries for feasible regions can be identified with distant functions [50]. For this reason generally the fitness function of a discrete CSP is based on violation counts as shown in Eq. (3). The solution of a CSP is $\vec{x} \in \{S_1 \cap S_2 \cap \ldots \cap S_m\}$ when all the constraints are satisfied. Generally weighted penalty functions like Eq. (4) are used to optimize a CSP, however, its main weakness is the difficulty to determine the appropriate weights when there is not enough information available for a given problem [19].

The main contribution of this paper is to show that exploitation of information derived from constraints using a generic operator leads to quality solutions. This paper is an extension of our work in [66] that solved discrete CSPs. We have enhanced the algorithm to solve discrete COPs as well. The paper is organized as follows: Section 2 describes the adaptation of intermarriage crossover operator for discrete search space. Section 3 extensively analyzes COPs for discrete search space. It describes how ICHEA has been incorporated to solve discrete COPs by adjustment of operators for optimization, search space analysis and resolving local optimal solution. This section also describes a generic fitness function applicable for a structured CP with different levels of strengths for constraints as an alternative to solve a new COP avoiding problem dependent penalty functions. Section 4 completes Section 3 by showing how the algorithm of ICHEA conforms to the changes made to solve a COP in a discrete search space. Section 5 shows experimental results of the benchmark exam timetabling problems. Section 6 discusses the experimental results and Section 7 concludes the paper by summarizing the results confirming the claim against the established hypothesis and proposing some further possible extensions to the research.

## 2   Intermarriage Crossover for Discrete CSPs

*Intermarriage* crossover for continuous search space defined in [58, 59] selects two parents from different *constraint satisfaction sets* to make them come closer iteratively towards their corresponding feasible boundary because the CSP solutions lie in the overlapping boundary region of feasible regions that satisfy different constraints. The iterative move for parent $P_i$ and $P_j$ to produce offspring $O_i$ is given as:

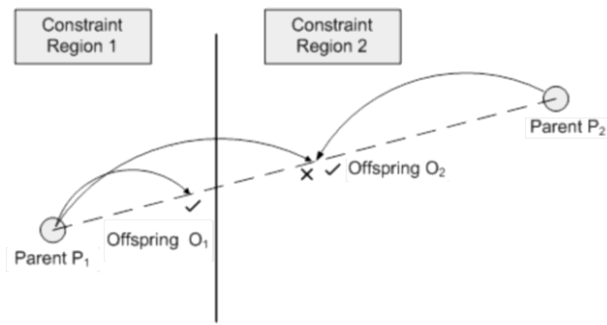

Fig. 1: Intermarriage crossover for continuous CSPs

$$O_i = r^i (P_j \oplus P_i) \qquad (5)$$

where $r$ is a coefficient in the range (0,1) which is generally 0.5. Variable $i$ gets incremented from 1 to a threshold value $T$ in the sequence $\langle 1, 2, \dots, T \rangle$. Operator $\oplus$ is a crossover between $P_i$ and $P_j$ which is minus ("−") for continuous search space. The intermarriage crossover process is shown in the Fig. 1 where ✓ mark indicates possible placement for an offspring and × mark indicate the offspring vector is unacceptable in that particular position. An offspring is accepted if it satisfies equal or more constraints than its corresponding parent. So using the Eq. (5) the next $i$ value is used until the offspring finds an acceptable place or a threshold value $T$ is reached. Favouring individuals that satisfy higher number of constraints and the use of feasible regions in *intermarriage* crossover guides the evolutionary search in finding the solution space quickly [58]. When the search space is discrete then the *intermarriage* crossover for continuous CSP cannot be used as it is to generate progenies as its formulation is different for continuous domain. The concept of *intermarriage* crossover is to fuse feasible solutions from two different constraint satisfaction sets together that makes the offspring "generic" that satisfy more constraints because its parents are from two different constraint satisfaction sets. The *intermarriage* crossover of two parents for discrete CSP transformed from Eq. (5) can be given as:

$$O_i = r^i(P_j \oplus P_i) = (P_j \oplus P_i) \qquad (6)$$

where operator $\oplus$ represents the fusion of two discrete feasible solutions. Here the value of coefficient $r$ and $i$ is 1 because fusion is non-iterative as shown in Fig. 2 where offspring are accepted if fusion results in better chromosome(s).

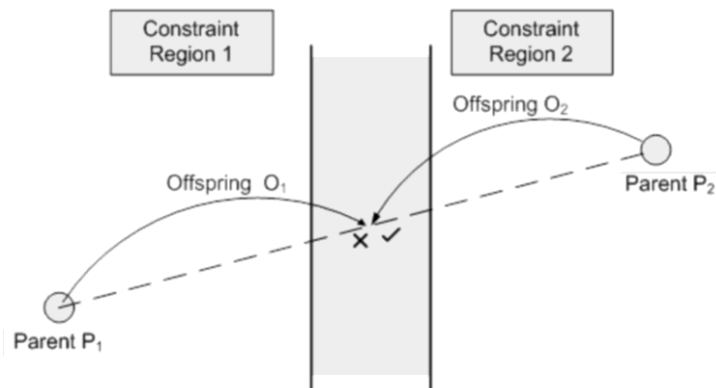

Fig. 2: Intermarriage crossover for discrete CSPs

ICHEA uses variable length chromosomes (partial solutions) to accomplish discrete valued *intermarriage* crossover where genotype is used as phenotypes. Variable length chromosome has been used in many applications as in [3, 60, 71]. Partial solutions are feasible solutions when only a subset of constraints has been considered. A CSP in $n$ dimensional search space has input vectors $\vec{x} = \{x_1, x_2, \ldots x_n\}$ with $m$ constraints. If only $k$ constraints have been satisfied or required to be satisfied then the partial solution $p = \{\vec{c_1}, \vec{c_2}, \ldots, \vec{c_k}\}$ represents a set of satisfied constraints $\vec{c_i}$ where $|\vec{c_i}| = n_i$ for $\forall i \in \{1, \ldots, m\}$ with $n_i \leq n$ and $k \leq m$. Each constraint can have different input variables $\{x_a, \ldots, x_b\}$ where $a, b \leq n$. Its cost and fitness function for a CSP can be given as:

$$cost(p) = m - |p| \qquad (7)$$

$$fitness(p) = |p| \qquad (8)$$

A partial solution $p_i$ takes part in *intermarriage* crossover by attempting to allocate constraints $\{\vec{c_a}, \ldots, \vec{c_b}\}$ from another partial solution $p_j$ into its existing feasible space (pattern of data points). To elaborate more we use an example of an N-Queen problem. N-Queen problem is a classic CSP that can be expressed as placing N queens on N x N chessboard such that no queen should be attacked by one another [37]. An N-Queen problem has one dimensional constraints $c_i = x_1$ for $\forall i \in \{1, ..., m\}$ where $i$ represents the column and the value of $x_1$ represents the row of the chess board. Suppose N-Queen problem of size 6 has two parents $P_1$ and $P_2$ with partial solutions $\langle 3, 6 \rangle$ and $\langle 6, 2, 5 \rangle$ respectively. The values represent constraints in the given order like parent $P_1$ satisfies two constraints where first value 3 represents first queen is at column 3 and first value 6 represents second queen is at column 6. The *intermarriage* crossover only tries to append/fuse the allele values of one parent into another one at a time until no more allele values is left. All the allele values that violate the constraints are dropped so the offspring are also feasible chromosomes. The generated offspring from these parents either satisfy equal or more constraints as shown in Fig. 3 where offspring $O_1: \langle 3, 6, 6 \rangle$ has two conflicting queens in row 6, offspring $O_1: \langle 3, 6, 5 \rangle$ has again two conflicting queens in row 6 and row 5 attacking each other diagonally. Offspring $O_1: \langle 3, 6, 2, 5 \rangle$ has no conflicts with the fusion. Each offspring has traits from both parents. An advantage of using variable length chromosome in this manner is reduction in computational time for N-Queen problem as shown in [66] where *intermarriage* crossover avoids recalculation of objective function because it only requires allele values to be appended. Its time complexity of *Big-O* order is $O(N)$ on average while traditional one-point crossover requires $O(N^2)$ on average as an N-Queen problem on chess board of size $N$ requires $N(N + 1)/2$ operations to check for constraint violations for each position of queens. One traditional one-point crossover for two offspring requires $N(N + 1)$ operations i.e. time complexity of $O(N^2)$. On the other hand, the *intermarriage* crossover only checks the violation of appended allele value with all other existing feasible values. It requires $(l_1 + l_2 + N) + N(l'_1 + l'_2)$ operations where $l_1$ and $l_2$ are the lengths of partial solutions of the parents and $l'_1$ and $l'_2$ are length of their respective non-duplicate allele values. The formulation of the time complexity is given in the Appendix. The first expression of time complexity $(l_1 + l_2 + N)$ indicates number of operations required to find the duplicate values. The second expression $N(l'_1 + l'_2)$ indicates the operations required to append the non-duplicate allele values to each other parents. The best time complexity is $(1 + 1 + N) + N(1 + 1) = 3N + 2$ operations when lengths of both parents are 1 and the worst time complexity is $(N/2 + N/2 + N) + N(N/2 + N/2) = N^2 + 2N$. As the evolutionary search progresses the length of partial solution increases towards maximum and length of non-duplicate allele values decreases to minimum. The generic time complexity can be written as $\big((N - \gamma) + (N - \gamma) + N\big) + N(d)$ where $\gamma$ is an integer value between $[1, N - 1]$ that decreases from $N - 1$ to 1, and $d$ is an integer value between $[2, N]$ that decreases from maximum to minimum non-duplicate value, as the evolutionary search progresses. ICHEA struggles and spent most of its computational time as the length of partial solutions increases which causes the constraints to become more intense. So the average computational activity occurs when the length of partial solutions increases that in turn decreases the length of non-duplicated allele values. In other words average time complexity is computed when $\gamma \to 1$ and $d \to 2$. The generic time complexity can be rewritten as $N(d - 2\gamma + 1)$ so $\lim_{d \to 2} \lim_{\gamma \to 1} N(d - 2\gamma + 1) = N$ as $d$ and $\gamma$ are both small integer values approaching towards respective minimum. Hence the average time complexity of *intermarriage* crossover has the *Big-O* order of only *O(N)*.

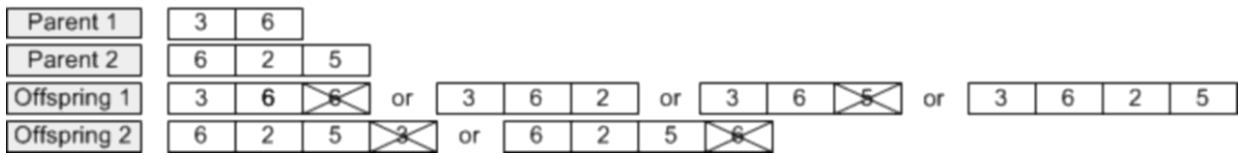

*Fig. 3: Variable length intermarriage crossover for N-Queen problem of size 6*

## 3    Solving Discrete COPs

So far we have seen that all the constraints must be satisfied to have an acceptable solution. Such constraints are known as *hard constraints*. Solutions, which satisfy all the hard constraints, are often called *feasible* solutions. In addition to the hard constraints there are usually various constraints that are considered to be desirable but not essential. These are often called *soft constraints* [7]. Soft constraints can have some degree of satisfaction or order of preferences for a particular problem. Soft constraints can be represented by penalty functions for COPs where higher weights demonstrate lower preferences and vice versa for higher preferences. However, the common problem of a penalty function as described in Section 1 is its dependency on the problem and difficulty in finding good weight factors like choosing or fine tuning number of parameters. Usually cost based weights are used in penalty functions which are problem dependent to cater for preferences of the constraints. For example, in a university timetabling problem, Prof. X might prefer teaching in the morning whereas Prof. Y prefers teaching in the afternoon [56]. The most basic form of a fitness function shown in Eq. (9) is given in terms of violations count where constraint strengths or degree of violation has not been considered [29, 30]:

$$f(v) = \sum_{p=0}^{D} v_p \qquad (9)$$

function $f(v)$ is the fitness value for $v$ satisfied constraints. $D$ is the lowest preference defined (highest numeric value for preference) which is one less than total number of preferences and $v_p$ is total number of satisfied constraints with $p_{th}$ preference. If a cost function is desired instead of fitness function then violation count $(L - v_p)$ can be used instead of $v_p$ where $L$ is the total constraints of a given problem. This is the most basic form of fitness function that only sums up the number of different constraint but does not rank them according to preferences. Eq. (9) can be modified to take preferences/strengths into account in the following function:

$$f(v) = \sum_{p=0}^{D} w_p v_p \tag{10}$$

where a problem dependent weight factor $w_p$ is used to give higher strengths to more preferred constraints.

In order to make a generic fitness function, the constraint strengths (preferences) and their relation with other constraints need to be considered. A constraint in a search space can have multiple degrees of constraint satisfaction relative to other constraints. A feasible region in a continuous search space is simply an overlapping region between constraints; however the overlapping region in a discrete search space can represent the degree of satisfaction from first preference to the last preference. Fig. 4 shows constraint *C1* has different degrees of satisfaction with other constraints. The *intersection* of *C1* with *C2* and *C3* results in constraint satisfaction with preference *p0*, constraint *C4* with preference *p3* and so on. If a generic fitness function is desired to maximize higher preferences then Eq. (11) ensures that constraints of higher preference are always given priority where $l_p$ is the total constraints satisfied with preference $p$ and $\beta$ is the upper bound for any given $l_p$. Fig. 4 has $l_0 = 2, l_1 = 0, l_2 = 0, l_3 = 1$ and so on. A generic value for $\beta$ is $L^2$, however any other problem dependent upper bound value can also be used in Eq. (11) if $(L^2 + 2)^{D-p}$ is too large to accommodate by a computer program. Eq. (11) also supports incrementality by reusing partial solutions that take consideration of higher preferences initially to build solutions that is followed by accommodation of remaining constraints of lower preferences. This equation gives higher fitness to a solution where constraints are solved with higher preferences.

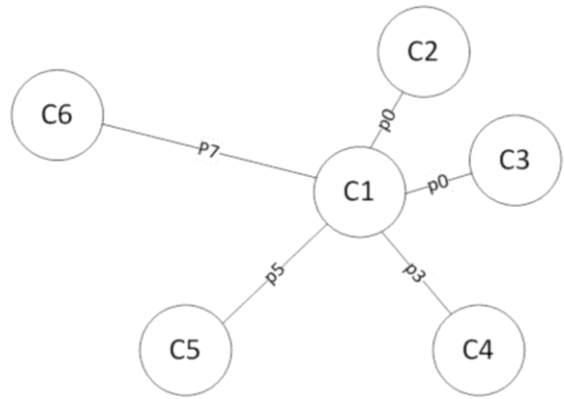

*Fig. 4: Constraint satisfaction with preferences in respect to other constraints*

$$f(l) = \sum_{p=0}^{D} l_p (\beta + 2)^{D-p} \qquad (11)$$

This equation encourages in maximizing $l_p$ for higher preferences, however, if the problem requires minimization of $l_p$ for lower preferences then the fitness function would be:

$$f(l) = \sum_{p=0}^{D} (L^2 - l_p)(L^2 + 2)^p \qquad (12)$$

Where we used $\beta = L^2$. For dynamic COP same equations Eq. (9) – Eq. (11) can be used with inclusion of parameter $t$ representing time or increment, consequently changing variable $l_p$ with function $l_p(t)$. Many large static COPs can also be solved incrementally by treating them as dynamic COP as discussed in Section 1 and in [57] where new constraints or a subset of constraints can be introduced into the search space after in every $g$ generations. This *divide and conquer* technique for solving a big COPs has shown better results than solving it otherwise in the experiments. A new fitness function given in Eq. (13) is derived to show the higher number of satisfied constraints should have higher fitness value. The proofs of Eq. (11), Eq. (12) and Eq. (13) are given in the Appendix.

$$f(l) = (L^2 + 2)^{D+1} \sum_{p=0}^{D} l_p + \sum_{p=0}^{D} (L^2 - l_p)(L^2 + 2)^p \qquad (13)$$

Note that these fitness functions are only applicable for single objective COPs and not for multi objective COPs. MOO uses *pareto* front to achieve the same results where decision makers can pick the results of their choice, not necessarily the ones where constraints are satisfied with mostly high preferences [27, 53].

So far couple of fitness functions are defined for discrete COPs but how these functions can be utilized in an EA or ICHEA is not discussed. Local search and hyper-heuristics are frequently used to solved these kinds of problems [11, 26, 56]. *Intermarriage* crossover described for discrete search space in Section 2 is only applicable for CSPs. For COPS some optimization techniques need to be incorporated. Commonly used crossover operators are generally not applicable as two feasible parents not necessarily produce feasible offspring and repairing infeasible offspring can be very expensive. So the only possible choice remain is to use the mutation strategies that look for neighbourhood solutions of feasible solutions in a population to search for the optimum solution.

### 3.1 Algorithms for discrete COPs

Our paper in [59] describes optimization techniques implemented in ICHEA for continuous COPs. This section explains *intermarriage* crossover together with additional mutation techniques commonly used for discrete COPs. The *intermarriage* crossover for discrete COP is influenced from our work in [59] using *influence* operator with Particle Swarm Optimization approach [28] where all swarm particles tend to move towards better positions nearby the best position that leads to optimum solution [28, 51]. This helps in exploring promising solution in a nearby region of the current best solution. If the *influence* operator is denoted by $\otimes$ then crossover between feasible solutions $P_i$ and $P_j$ involves the following steps:

1) $P_i' = P_i \otimes P_j$

2) $P_j' = P_j \otimes P_i$

3) $P_i' = P_i' \otimes P_{best}$

4) $P_j' = P_j' \otimes P_{best}$

The influence operator simply tries to influence chromosome $A$ with predefined number of allele value(s) of chromosome $B$ (called degree of influence) as shown in Fig. 5. Suppose chromosome $A = \{4, 2, 5, 1, 3\}$ is influenced with second allele value of chromosome $B = \{3, 1, 4, 2, 5\}$ that moves the influenced value 1 at second position from current fourth position in chromosome $A$. If a move is not valid then it can be repaired, rejected or kept separately in the repertoire of infeasible solutions which can be persisted with influence operator until a feasible solution is obtained. Influence operator brings one chromosome closer to another.

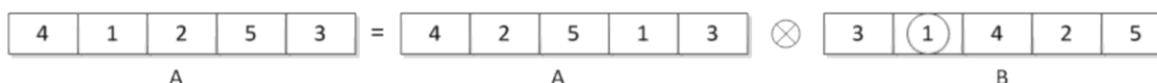

*Fig. 5: Influence Operator: Chromosome A is influenced with the second allele value of Chromosome B*

In addition to the crossover some mutation strategies can also be applied like swapping two allele values or group of allele values, or remove some allele values and place them in a different position. Many times these operators are problem dependent to define their feasibility after the move. For example in exam timetabling problems experimented in this paper we defined the following operators to search for an optimum solution.

1. Mutation by traditional *Kempe* chain: the *Kempe* chain has been primarily used in graph coloring problem but it has been proven successful in timetabling problems as well [11, 26, 69]. It starts with moving $1 - 5$ elements randomly picked from timeslot *I* to *J* but this may cause conflicts in timeslot *J* so all conflicting elements from *J* is moved to *I* but again the inclusion of new elements from *J* may cause conflict in *I*. So all conflicting elements are moved back and forth from *I* to *J* until no conflicts remain.

2. Mutation by boundary *Kempe* chain: we also propose a variant of traditional *Kempe* chain for exam timetabling problem called boundary *Kempe* chain where selection of timeslot *I* is based on locations of most conflicting elements of the timetable and timeslot *J* is selected from far ends of the timetable as it is likely to have less conflict on the ends because of less timeslots at one end. Timeslot *I* is randomly picked from the locations of top 10% of the most conflicting elements (exams) and timeslot *J* is selected from the farthest or next to farthest timeslots either from beginning or end of the timetable.

3. Swap two elements: randomly swap two elements (exams) which may result in an infeasible solution. These infeasible solutions are kept in a separate set which is later repaired using traditional Kempe chain by removing the infeasible element from timeslot *I* to *J*.

*4.* Swap the whole group (time slot): either swap two timeslots or move a timeslot from one place to another randomly.

5. Mutation by removal: remove an element (exam) randomly then try to insert it into a different timeslot. Reject the solution if infeasible.

6. Mutation cluster: randomly pick and add an exam from a cluster of possible exams that a timeslot can have for a feasible solution which is not already in that timeslot. Locate this exam that is elsewhere in the timetable and delete it.

7. Crossover using influence operator [59]: this can be used between feasible and infeasible chromosomes described earlier. A predefined number of infeasible solutions neighboring a feasible solution called a *community* move closer towards a feasible solution that can transform infeasible solutions into a feasible solution. Community members do not move to other communities while they are infeasible however, only the best feasible member within a community influences other members. The major drawback here is infeasible solution can become a duplicate of the feasible solution.

8. Crossover using *Kempe* chain: as described earlier the *intermarriage* crossover between two feasible solutions can be performed using influence operator; however, influence operator might produce an infeasible solution because of its *move*. The *move* of *Kempe* chain can be used that produces feasible solutions, instead of the influence operator's *move* for exam timetabling problem. This is done by first locating the timeslot carrying the selected allele value of the influencer chromosome *B* to influence the other chromosome *A*. We record the position of this timeslot in chromosome *B*. The same position of a timeslot in chromosome *A* works as timeslot *I* for chromosome *A*. Then we locate the timeslot carrying the same allele value in chromosome *A* which will be designated as timeslot *J*. Now we apply traditional *kempe* chain on chromosome *A*. This would influence the chromosome A with chromosome *B* and at the same time produce the feasibility intact. The process of selecting *I* and *J* for *kempe* chain *move* has been also shown in Fig. 6 where allele value $e_i$ from chromosome *B* influences chromosome *A*. Timeslot *I* and timeslot *J* is determined by the positions of the column containing $e_i$ in chromosome *B* and chromosome *A* respectively. Once timeslot *I* and timeslot *J* are found *kempe* chain

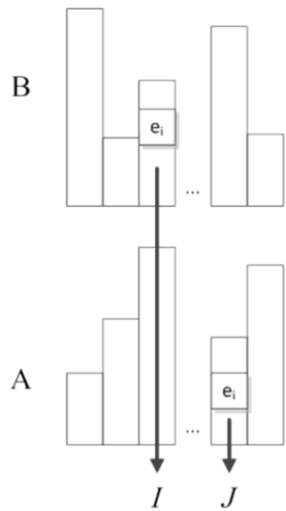

*Fig. 6: Crossover using kempe chain for influence operator*

mutation is applied on chromosome A.

Each of these operators runs in a sequence in ICHEA where next operator is selected when there is no improvement for the best so far solution for *s* generations. Stagnant generations *s* is 5 in our experiment as higher value slows down the process and getting another operator is likely to resolve the stagnant state. In hyper-heuristic domain these operators can be used as lower level heuristics where sequence of operators is chosen heuristically. This is a cyclic process until the termination condition is met.

Local search using the above mentioned operators can be useful to maintain the feasible solutions and search for optimal solution through neighborhood search. Hill-climbing algorithm is a canonical form of a local search, however, the basic hill-climbing algorithm generally provides locally optimum values which depends on the selection of the starting point. If lucky then some of these initial locations will have a path that leads to the global optimum [44]. Hill-climbing is an iterative process where each candidate solution searches in its nearby region for better solutions. This becomes even more difficult when the search space is discontinuous with constraints. More sophisticated local search techniques are Simulated Annealing (probabilistic hill-climbing) and Tabu Search where they apply some mechanism to escape from local optimal solution [11, 44].

We propose a new local search strategy which is called *revertible clonal hill-climbing* (RCHC). The closest resemblance of RCHC is hill-climbing with *tabu* search. However it differs in management of *tabu* list and inclusion of backtracking mechanism. Firstly feasible solutions $pop_{COP}$ from the whole population $pop$ are retrieved through intermarriage crossover describe in Section 2 where each feasible solution of the population will carry a set of *tabu* list and a set of previous solutions. The procedure begins by creating number of clones for each individuals inspired by CLONAX [61] algorithm where each clone goes through allocated mutation process. The size of the clones is determined by a formula $N_c = min\left(5, \sum_{i=1}^{|pop_{COP}|} \frac{\alpha|pop|}{i}\right)$ similarly used in [18] for CLONALG algorithm where $N_c$ represents size of the clones for a partial solution $i$ sorted from best to worst. $\alpha$ is 1 in our experiments. This formula creates more clones for better solutions. A mutant clone immediately replaces the incumbent solution if it has better fitness value. If an individual has not been improved in $t$ generations then it is reverted to its previous state and the current state is marked *tabu*. The algorithm keeps the history of $H$ changes which is generally in the range of $[0, 5]$. This is similar to the *backtrack* algorithm when a solution is not improving for $\delta$ generations then this solution is reverted to its parent branch and the new search is started so that poor solution can be eliminated with the new solutions to promote diversity. However this algorithm has limited number of backtracks for efficiency purpose. The pseudocode of RCHC is given in Fig. 7. The size of *tabu* list and previous states are constant. If the lists are full then the new record replaces the oldest record. If all previous solutions are retrieved then revert process removes the individual from the population and new feasible solution is added into $pop_{COP}$.

Fig. 8 shows backtracking steps of a feasible solutions starting at position $a$. When the solution is improved it traverses through positions $a \rightarrow b \rightarrow c \rightarrow d$ iteratively. When position $d$ does not progress it reverts to position $c$ which is then improved through mutation and finds a new position $e$. After some time position $e$ becomes stagnant and similarly it traverses through $e \rightarrow c \rightarrow b \rightarrow f \rightarrow g \rightarrow h$ where $h$ is its latest position. Iterative steps are numbered in sequence in the figure.

```
i ← 0
Each vector V_i of the population of
feasible solutions {V_i} has its
corresponding stack H_i to store previous
changes to the solutions and
corresponding queue T_i to hold tabu list.
For each vector V_i of the population
{C} = clone(V_i)
  k ← 0
  For each clone of V_i
    C_k ← mutation(C_k)
    If f(C_k) > f(V_i) Then
      Push the vector V_i in its
      corresponding fixed sized
      stack  H_i ← Push(H_i, V_i)
      V_i ← C_k
    Else
      If V_i is not progressive Then
        set V_i as tabu by enqueuing it
        into its  corresponding
        fixed sized queue
        T_i ← Enqueue(T_i, f(V_i))
        Revert V_i to its previous solution
        V_i ← Pop(H_i)
        Terminate For loop
      End if
    End If
    k ← k + 1
  End For loop
  i ← i + 1
End For loop
```

*Fig. 7: Pseudocode for revertible clonal hill-climbing*

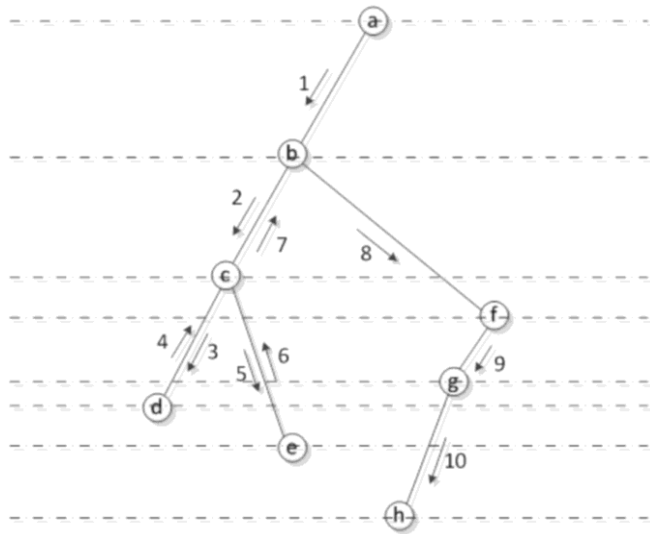

*Fig. 8: Revertible hill-climbing for a feasible solution*

## 3.2 Search space analysis

Population based meta-heuristic like EAs use greedy approach by keeping the good solutions and discarding solutions with low fitness values, however, these low fitness individuals can be promising depending on their locality in the search space. Some basic types of convergence plot for local search and their characteristics can be seen in Fig. 9 where diagram 1 shows a small and steep hill, diagram 2 shows a steep hill with larger spread, diagram 3 has small elevations (local optimum), diagram 4 has a gradual elevation on a bigger space, diagram 5 has a little elevation followed by a big plateau which elevates again and diagram 6 is same as diagram 5 but with a hole (constraint) in the plateau. More possibilities of the convergence plots are possible but the objective here is to demonstrate that a point with lower fitness value in a search space can lead to the global optimum solution. If the population of EA is not managed properly then greedy approach tends to discard promising points in early generations. For example if diagram 4 leads to the global optimum and EA starts with points that have higher fitness values located near the local optimum regions of diagram 1 or 2 together with some point in the region of diagram 4 towards the lower elevation. As the algorithm progresses the points of diagram 4 may not survive because of their lower fitness value. Once these points are deleted local search through neighborhood would fail to locate the neighborhood of global optimal solution.

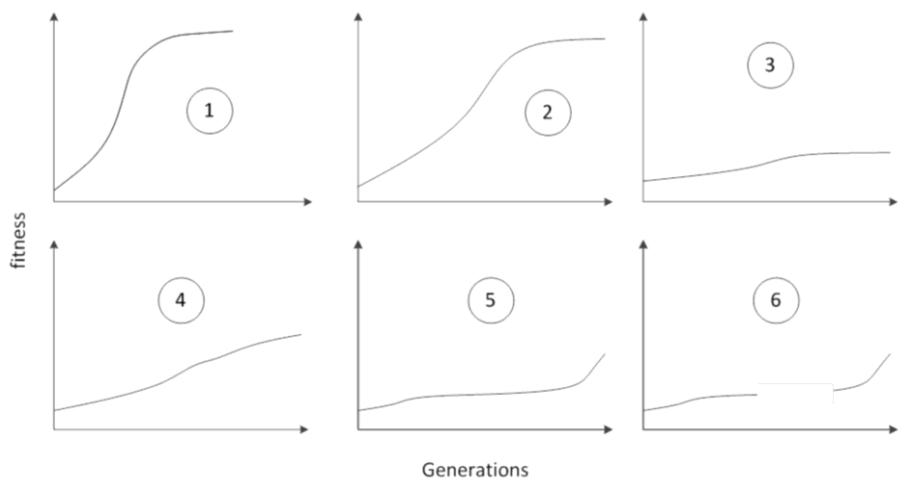

*Fig. 9: Types of convergence plots in a search space*

ICHEA is a population based EA that uses RCHC, a local search strategy for discrete COPs. Given the scenario above ICHEA does not filter out solutions based on fitness value alone. If a particular point has low fitness value but its fitness value gradually increases then it is considered promising without comparing it with other individuals of higher fitness value. When a solution becomes stagnant then it is reverted or deleted as described in RCHC algorithm.

### 3.3 Stalled local optimal solutions management

It has been noted through experiments that ICHEA still gets stalled in local optimal solution. We applied a similar strategy of *tabu* search algorithm based technique described for continuous COPs in our previous work in [57] which is further enhanced to suit the needs of discrete search space. We keep the history of $t$ previous best solution achieved so far to determine the *tabu* region of the search space using the following formulation:

$$S_{tabu_1} = S_{curBest} \cap S_{prevBest_1}$$

$$S_{tabu_{i+1}} = S_{tabu_i} \cap S_{prevBest_i} \; for \; \forall \; i \; \in \{2, 3, ..t\} \tag{14}$$

Eq. (14) shows the *tabu* region in the search space by $S_{tabu_i}$ where $i \in \{1, 2, 3, ..t\}$ for a predefined constant value $t$ which is generally 5. For discrete search space $S_{tabu_i}$ can be just a sequence/pattern of allele values. $S_{curBest}$ is the best solution of the current generation and $S_{prevBest_i}$ is the previous $i^{th}$ best solution. The intersection of $S_{tabu_i}$ and $S_{prevBest_i}$ retrieves the common or unchanged allele values from either current best to previous best or current *tabu* region to previous best. This is done to track the traversal of the best solution and try to divert the whole population that is stalled in the local optimal region. If $S_{tabu_i}$ for $i = 1$ does not improve the best solution then $i$ is incremented to get the next *tabu* region. Fig. 10 shows the incremental shrinking of the *tabu* region. Part (A) shows the traversal of the best solution, Part (B) and Part (C) show the reduction of the size of *tabu* region for next increments based on Eq. (14). As the *tabu* region decreases in size it is more likely to find chromosomes that have similar pattern to the *tabu* region. In other words when a *tabu* region becomes subset of a chromosome in terms of pattern matching of allele values then the incumbent chromosome is said to be a *tabu* which is shown in Eq. (15) where $\omega(S_i)$ is a function denotes if a chromosome $S_i$ is tabu or not. For example a pattern of *tabu* region can be denoted as $S_{tabu} = \langle 2, 5, \#, 1, 3, \# \rangle$ where # indicates a chromosome in question can have any value at that position. A chromosome $S = \langle 2, 5, 4, 1, 3, 6 \rangle$ is a subset of $S_{tabu}$ but $S = \langle 2, 5, 1, 4, 3, 6 \rangle$ is not.

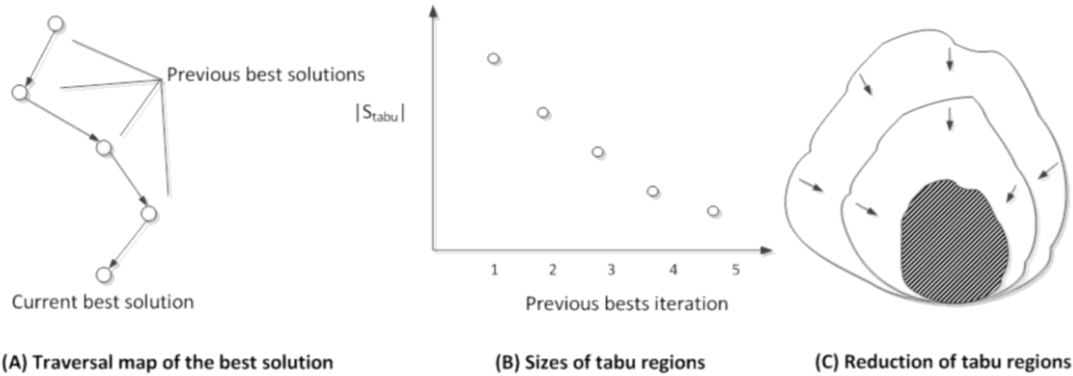

(A) Traversal map of the best solution  (B) Sizes of tabu regions  (C) Reduction of tabu regions

Fig. 10 *Shrinking of tabu region for stalled ICHEA*

$$\omega(S_i) = \begin{cases} 0, & if\ S_i \subseteq S_{tabu_i} = \emptyset \\ 1, & if\ S_i \subseteq S_{tabu_i} \neq \emptyset \end{cases} \quad (15)$$

## 4   ICHEA Algorithm on Discrete Search Space

Some CPs like exam timetabling problems has many constraints. These constraints can be divided into several components (subsets of constraints) then each component can be solved incrementally. This divide and conquer approach solves a CP by taking each component in turn to get feasible partial solutions. Partial solutions are solutions which satisfy all the constraints present at a given point in time in a search space. Solving CPs incrementally has many advantages. It also comes handy when a new constraint is added or an existing constraint is changed. It is also useful in doing what-if analysis on strengths of constraints or inclusion/exclusion of constraints, hence supporting exploratory nature of searching for various solutions. A by-product of incrementality based search is a set of generated partial solutions for each increment that can be stored separately and later reused, where a new constraint can be added or an existing constraint can be changed without making too much distortion to the current solution. Suppose there are $n$ sets of partial solutions $\{P_1, ..., P_n\}$ where each partial solution $P_i$ satisfy $L_i$ constraints and carries $N_i$ feasible solutions. Fig. 11 shows the algorithm for reusing partial solutions to cater for any changes to the constraints. The graphical interpretation of the algorithm is given as Fig. 12 where a constraint $C_x$ is presented to partial solutions. It is assumed that constraint $C_x$ has the lowest strength. Partial solutions $P_4$ and $P_3$ are unable to accommodate $C_x$ but $P_2$ gives feasible solutions with $C_x$. Now the previous partial constraints are no longer valid which are replaced by $P'_2$, $P'_3$ and $P'_4$ recpectively. All these new partial solutions satisfy the given constraint $C_x$. The algorithm describes the addition of a new set of constraints $C$

```
i ← n
While (i ≥ 1)
  P_i ∈ S_i^{N_i×L_i}  % the structure of P_i can be
  kept intact if no distortion of the
  solution is allowed.
  Run ICHEA on existing partial solutions
  P_i and add a set of given constraints
  C for the feasibility check for a given
  number of generations.
  % Define parents pool size d which is
  generally T/2
  T = {P_i ∪ C}  % T is total population
  For Each Generation
    X ← Selection(T);
    P'_i = {p ∈ X | X ⊂ T ∧ |X| = d}
    % Intermarriage crossover using
    operator ⊗.
    P''_i^j ← p1 ⊗ p2 where p1 ∈ P'_i and p2 ∈ P'_i
    and p1 ≠ p2
    P''_i = {∀P''_i^j | j ∈ {1,…,N_i}}.
    Mutation(P''_i);
    X = {P''_i ∪ T}  % Collect all chromsomes
    T ← SortAndReplace(X);
    CheckTerminationCriteria();
  End for loop;
  Solution is found when {∃T^j ∈ T | {C ∪ P_i^j} −
  T^j = ∅ ∧ j ∈ {1,…,|T|}} however, feasible
  solutions can be stored for optimization
  purpose in T''.
  T'' = {∀T^j ∈ T | {C ∪ P_i^j} − T^j = ∅ ∧ j ∈ {1,…,|T|}}.
  If (T'' ≠ ∅) Then
    Print("solution is found")
    Print("Input T'' into ICHEA to get full
    solution");
    Terminate While loop
  Else
    If P_i has same or higher strength (≽)
    than C then we need to keep all
    constraints of P_i.
    If (P_i ≽ C) Then
      F = {∀c ∈ {T^j − P_i^j} | P_i^j ⊂ T^j ∧ j ∈ {1,…,N_i}}
    Else
      F = {∀c ∈ {T^j − C} | C ⊂ T^j ∧ j ∈ {1,…,N_i}}
    End if
    Print("List of combinations of
    constraints not solved:")
    Print(F);
    Print("Moving to next partial
    solution");
    % Stop if user decides.
    i ← i − 1
  End if
End While loop
```

*Fig. 11 Reusing partial solutions in ICHEA for new addition of constraints*

that is verified against partial solutions to get feasible solutions. If a current partial solution is unable to accommodate constraints $C$ then these constraints are tested with previous partial solutions iteratively until all the partial solutions are exhausted. If constraints are structured according to their respective strengths then the algorithm is biased towards retaining solutions that solves more constraints of higher strengths rather than keeping the existing partial solutions. We used the following notations for the algorithm:

— $X_i^{R_i \times C_i}$ : Matrix X on $i^{th}$ increment has $R_i$ rows and $C_i$ columns.

— $P_i$: Set $P_i$ on $i^{th}$ increment.

— $P_i^j$: Vector j of set $P_i$ on $i^{th}$ increment.

— $\otimes$: Intermarriage crossover operator

— %: Additional comments afterwards

This algorithm is useful in doing what-if analysis or updating constraints by revisiting previously solved partial solutions. If the current solution is to be kept intact then the allele values in the chromosomes can be locked to prevent any changes through evolutionary operators. However, the search space becomes more constrained when the problem is required to be optimized. Any changes in an existing constraint can also be backtracked to the increment where it was resolved. For real time DCPs this algorithm is also very helpful as it does not require restarting the whole process unless none of the partial solutions are able to accommodate the changes to the constraints.

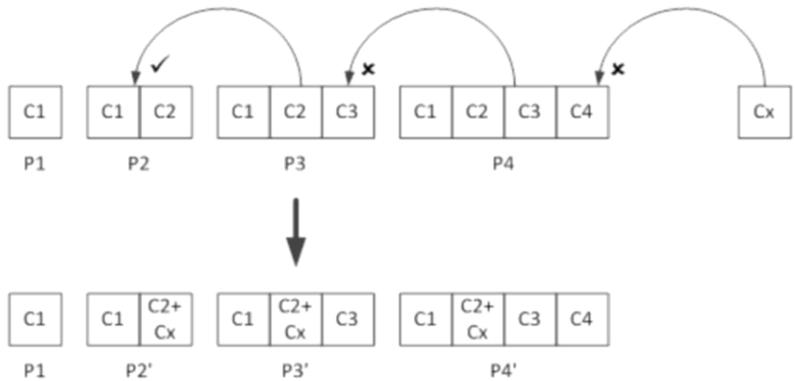

*Fig. 12 Incremental process shown diagrammatically*

For continuous CPs in [57] each component consists of one constraint only which were sorted decreasingly based on their constraint strengths $\rho$. In the literature, exam timetabling problems sort the constraints according to the largest degree (LD), saturation degree (SD), largest weighted degree (LWD), largest penalty (LP) or random Order (RO) [8, 14, 15]. LD and SD are commonly used sorting order. In LD exams are ordered decreasingly according to the number of conflicts each exam has with others, and in SD the exams are ordered increasingly according to the number of remaining timeslots available to assign them without causing conflicts. The definition of other sorting orders can be found in [14]. ICHEA uses LD to sort all the exams based on clashes with other exams. It takes only 5% of the sorted exams in every increment and once a feasible solution is obtained the optimization operators are applied for $G$ generations before taking next increment of the exams (constraints). The value of $G$ is 50 in our experiments. Apart from the addition of incremenatily feature for what-if analysis the structure of ICHEA is not changed. As discussed in [59] ICHEA runs two processes in parallel – one to solve CSP and another to optimize CSP solutions. The parallel process starts by dividing the whole population $pop$ into 2 parts. First part $pop_{COP}$ keeps the feasible solutions that are required for optimization and the second part $pop_{CSP}$ keeps the *good* infeasible solutions that are processed to get CSP solutions. The ratio of $pop_{COP}:pop_{CSP}$ is 1:1 for our experiments. The pseudocode of ICHEA is given in Fig. 13 followed by the description. Initialization of chromosomes and operators for reproduction has been modified but the structure of ICHEA is still same as any other EA.

```
chromosomes  = initializeChromosomes();
for each generation
        parents = Selection();
        offspring =
        interMarriageCrossover(parents);
        Mutation(offspring);
        chromosomes = chromosomes +
        offspring;
        SortAndReplace();
        CheckTerminationCriteria();
End for loop
```

*Fig. 13 Pseudocode for ICHEA*

**InitializeChromosomes:** ICHEA promotes incrementality and it can solve a given CP in an incremental manner. To solve a discrete CP incrementally, only a set of constraints is considered at a time. For example if there are total of $m$ constraints for a static CP then only $r$ proportion can be considered at a time for certain number of generations. A population of chromosomes for the first increment can be randomly generated using sequence of integer values $\langle 1, 2, \ldots rm \rangle$ where each number represents a constraint. For next $(i + 1)$ increment the new chromosomes have values in the range $\langle irm + 1, irm + 2, \ldots (i + 1)rm \rangle$. If $(i + 1)rm > m$ then it is turned into $m$.

**InterMarriageCrossover:** The crossover techniques have been described in Section 2 works well for discrete CSPs but it is not applicable for COPs. COPs depend on mutation strategies only as general crossover techniques mostly produce infeasible solutions which would require a separate repair technique to transform all the infeasible offspring to feasible offspring which is normally computationally expensive [20].

**Mutation:** Many mutation strategies can also produce infeasible solutions but it can be less expensive to repair compared to offspring produced through common crossover operators. Additionally it suits well for local search techniques. As mentioned in Section 3.1 ICHEA uses RCHC to optimize a CP. Once a chromosome solves all $irm$ constraint for $i^{\text{th}}$ increment, it is considered a partial solution for COP which is then qualified to apply aforementioned mutation strategies and *influence* operator to optimize the exiting partial/full solutions. *Influence* operator is particular useful for CP as this operator moves infeasible solutions towards feasible solutions.

**SortAndReplace:** sorting is not required for feasible solutions ($pop_{COP}$) as RCHC only keeps the good solutions and poor solution are discarded. Population of infeasible solutions ($pop_{CSP}$) are increased by $\kappa$ additional solutions due to random selection of parents in *intermarriage* crossover operator so an intelligent sorting is required to keep the good and diverse solutions because as the generation progresses EAs tend to preserve better chromosomes and poor chromosomes die away. If only best ones are kept then diversity is lost [34]. We first sort the whole $pop_{CSP}$ then pick the solutions using the following exponential function in Eq. (16) that keeps the good solutions as well as maintains the diversity. The size of the whole population remains intact. The details of the Eq. (16) is given in the Appendix.

$$f''(i) = \begin{cases} f'(i-1) + 1, & \text{if } f'(i) \leq f'(i-1) \\ f'(i), & \text{otherwise} \end{cases} \quad (16)$$

where

$$f'(i) = \left\lfloor (|POP_{CSP}| + \kappa) \frac{f(i) - f(1)}{1 - f(1)} \right\rfloor \text{ and}$$

$$f(i) = e^{-\rho \left( \frac{|POP_{CSP}| - i}{|POP_{CSP}|} \right)} \text{ with value of } \rho = 5.$$

Other functionalities of ICHEA are same as defined in [58] for continuous search space.

# 5 Experiments

A good example of a discrete COP is a university exam time tabling problem that has been attracting the attention of the scientific research community across Artificial Intelligence and Operational Research for more than 40 years [8, 54]. A timetable in general is a placement of a set of meetings in time. A meeting is a combination of resources (e.g. rooms, people and items of equipment). An instance of a timetabling problem is university exam timetabling where exams are required to be spread out sufficiently for all the students to give them break so they are better prepared for next exams. Exams must be scheduled so that no student has more than one exam at a time [10].

A common focus in the literature has been mainly to produce optimum solution with lowest cost function indicating high spread of exams for each student [10, 11, 26, 39]. Efficiency is an important aspect of optimization problems, however conforming to a time limit is not an important constraint in real world timetabling [1, 26]. An important but generally over-sighted issue with timetabling problem is regeneration of a timetable using previously prepared timetables. The traditional techniques in any university require lots of user input where a lot of work from the technicians is required every time a new time table is generated. The drawback of EAs of not imitating real human behavior of learning from past can be labeled as an "unintelligent artificial intelligence" solution. A simple solution to this problem is to provide a set of partial solutions that are solved incrementally according to their preferences and which can later be combined to get the final solution. Some partial solutions can also be reused, updated or removed and then merged again to get the final solution more efficiently without regenerating the whole solution.

As discussed in Section 3 there are two ways to provide fitness function for a CP: a problem dependent weight based penalty function and a generic penalty function. We used University of Toronto benchmark exam timetabling problems (version I) given in [54, 73] where the given weights based on the spread of exams for each student is:

$$W_d = S_d 2^{4-d} \tag{17}$$

where $d$ is the distance between two timeslots in the range [0 4], $S_d$ is total corresponding students and $W_d$ is the total corresponding weight. The cost function is the average weight corresponds to each student given as:

$$f = \frac{1}{S}\sum_{d=0}^{4} S_p 2^{4-d} \tag{18}$$

However, Eq. (13) can be used for generic penalty function where $l_p$ indicates total exams violating constraints of $p^{th}$ preference. We mostly focused on standard weights based fitness function for benchmark exam timetabling problem as there are lots of published results to compare with. At the end we only had limited experimental results for generic fitness function since there are no known results to be compared from our knowledge. Currently, the generic fitness functions given in Eq. (11) – Eq. (13) have a major drawback that it is not able to handle overlapping (mutually inclusive) constraints like maximal spread of the exams for each student.

**5.1 Problem dependent weights based fitness function**

Hyper-heuristics have been frequently used to solve benchmark exam timetabling problems which show promising results. The same hyper-heuristics can be applied to the same class of problems like graph coloring problem and exam timetabling problems [11, 26]. Hyper-heuristics heuristically selects these mutation strategies that best suits a given problem. ICHEA is a meta-heuristic algorithm that uses multiple mutation strategies to optimize a CP as described in Section 3.1. All the benchmark problems have been experimented on a Windows 7 machine with Pentium (R) i5 CPU 2.52 GHz and 3.24 GB RAM except the problem Pur93 which was run on a server machine (Intel Xeon CPU 2.90GHz and 128 GB RAM) because of its size and memory requirements. No parallel processing or distributed environment has been used for the experiments. We ran all the problems over-night because of their size and complexity. Additionally, real world timetabling problem does not required to be solved within minutes or hours [1, 26]. Even though smaller sized problems like Hec92 and Sta83 can be solved within an hour or two; however problem Pur93 had to be run for almost 24 hours because of its huge size. All the experimental results have been verified through the standard evaluator program available in the dedicated website for research on benchmark exam timetabling problems [73].

*TABLE I Parameter settings for ICHEA/IICHEA for benchmark exam timetabling problems*

| Parameter | Value |
|---|---|
| Population size | 100 |
| Optimizing partial solution for $G$ generations in each increment | 50 |
| Degree of influence | 3 |
| Community size | 4 |
| Total communities | 10 |
| Stagnant generations | 5 |
| A constant $\beta$ to generate clones | 1 |
| Set of $H$ changes for a solution in RCHC | 3 |
| Stagnant for $\delta$ generations to start backtrack | 5 |
| History of $t$ previous best solution for a tabu set | 5 |

As discussed in [57] ICHEA is able to incrementally solve a dynamic CP. Many times incremental approach to solve a complex static CP gives better results than solving entire constraints altogether. We used both approaches in the experiments to demonstrate supremacy of one approach over another. We observed that this incremental approach also helps in quickly providing feasible partial solutions and eventually feasible solutions at the success rate (SR) of 100% for all the benchmark problems. SR is the rate of successful trials for each problem i.e. $SR = successful\ trials\ /total\ trials$. SRs of non-incremental ICHEA are very low for bigger problems like Car91 and Uta92 have only 0%-10% of SR, and 30%-70% for other problems of medium size. Non-incremental ICHEA also takes much longer duration to get the first feasible solution. The unpromising outcome from non-incremental ICHEA has led us to do the experiments with incremental ICHEA only. We first sort the constraints (exams clashes) according to LD then remove first 5% of the total exams as input for each increment in ICHEA. *Intermarriage* crossover constructs new partial feasible solutions which are then optimized using mutation strategies for feasible partial solutions. We used two instances of ICHEA for the experiments to demonstrate the validation of incrementality. The first and second instances of ICHEA optimize the partial solutions for 0 and 50 generations respectively. The only difference between these two instances is the first one does not apply optimization strategies to partial solutions while the other optimizes the partial solution for 50 generations. However, both instances get the feasible solutions incrementally. To distinguish the two instances the first one is called ICHEA and second one is called incremental ICHEA (IICHEA) as it fully exploits the notion of incrementality. These partial solutions consist of exams from current and all previous increments being allocated in the timetable. The total available timeslots of the exam timetables are always fixed to the given value. Constraint optimization in ICHEA is a parallel process of finding feasible partial solutions from infeasible partial solutions, and optimizing feasible partial solutions as discussed in Section 4. This has also been realized previously with experimental results on continuous domain in [57]. More importantly ICHEA does not have to define any problem specific algorithm to get feasible solutions as many other approaches like [1, 13, 26] use bespoke algorithms or SD graph-coloring heuristics to get the feasible solutions.

*TABLE II Statistical summary of results from IICHEA and ICHEA*

| Instance | Best | Median | Worst | SD |
|---|---|---|---|---|
| Car91 | 4.91 (5.1) | 5.04 (5.3) | 5.16 (5.46) | 0.01 (0.15) |
| Car92 | 4.08 (4.3) | 4.1 (4.45) | 4.2 (4.54) | 0.05 (0.10) |
| Ear83 | 33.24 (33.6) | 34.02 (34.69) | 34.7 (37.39) | 0.57 (1.24) |
| Hec92 | 10.13 (10.17) | 10.33 (10.45) | 10.61 (11.15) | 0.15 (0.37) |
| Kfu93 | 13.58 (13.8) | 13.8 (14.1) | 14.21 (15.09) | 0.20 (0.37) |
| Lse91 | 10.37 (10.95) | 10.51 (11.34) | 10.67 (11.8) | 0.11 (0.27) |
| Pur93 | 4.67 (5.2) | 4.78 (5.43) | 4.99 (5.81) | 0.12 (0.21) |
| Rye92 | 8.63 (9.07) | 8.76 (9.4) | 8.85 (9.7) | 0.08 (0.19) |
| Sta83 | 157.03 (157.03) | 157.03 (157.03) | 157.03 (157.03) | 0.0 (0.0) |
| Tre92 | 8.33 (8.8) | 8.5 (9.3) | 8.8 (9.6) | 0.16 (0.28) |
| Uta92 | 3.28 (3.48) | 3.41 (3.60) | 3.57 (3.64) | 0.07 (0.07) |
| Ute92 | 24.85 (24.9) | 24.9 (25.7) | 25.1 (27.0) | 0.10 (0.87) |
| Yor83 | 36.24 (36.45) | 36.6 (37.04) | 38.8 (39.69) | 0.71 (1.15) |

*TABLE III Best results from the literature compared with IICHEA*

| Algorithms | Car91 | Car92 | Ear83 | Hec92 | Kfu93 | Lse91 | Pu93 | Rye92 | Sta83 | Tre92 | Uta92 | Ute92 | Yor83 |
|---|---|---|---|---|---|---|---|---|---|---|---|---|---|
| IICHEA | 4.9 | 4.1 | 33.2 | 10.1 | 13.6 | 10.4 | 4.7 | 8.6 | 157.0 | 8.3 | 3.3 | 24.8 | 36.2 |
| Carter et al. [16] | 7.1 | 6.2 | 36.4 | 10.8 | 14.0 | 10.5 | 3.9 | 7.3 | 161.5 | 9.6 | 3.5 | 25.8 | 41.7 |
| Merlot et al. [44] | 5.1 | 4.3 | 35.1 | 10.6 | 13.5 | 10.5 | - | 8.4 | 157.3 | 8.4 | 3.5 | 25.1 | 37.4 |
| Casey and Thompson [17] | 5.4 | 4.4 | 34.8 | 10.8 | 14.1 | 14.7 | - | - | 134.9? | 8.7 | - | 25.4 | 37.5 |
| Yang and Petrovic [78] | 4.5 | 3.9 | 33.7 | 10.8 | 13.8 | 10.4 | - | 8.5 | 158.4 | 7.9 | 3.1 | 25.4 | 36.4 |
| Abdullah et al. [1] | 5.2 | 4.4 | 34.9 | 10.3 | 13.5 | 10.2 | - | 8.7 | 159.2 | 8.4 | 3.6 | 26.0 | 36.2 |
| Eley [33] | 5.2 | 4.3 | 36.8 | 11.1 | 14.5 | 11.3 | - | 9.8 | 157.3 | 8.6 | 3.5 | 26.4 | 39.4 |
| Burke and Bykov [6] | 4.6 | 3.8 | 32.7 | 10.1 | 12.8 | 9.9 | 4.3 | 7.9 | 157.0 | 7.7 | 3.2 | 27.8 | 34.8 |
| Burke, Eckersley, et al. [11] | 4.9 | 4.1 | 33.2 | 10.3 | 13.2 | 10.4 | - | - | 156.9 | 8.3 | 3.3 | 24.9 | 36.3 |
| Demeester et al. [27] | 4.5 | 3.8 | 32.5 | 10.0 | 12.9 | 10.0 | 5.7 | 8.1 | 157.0 | 7.7 | 3.1 | 24.8 | 34.6 |

The parameter settings for IICHEA and ICHEA to solve the benchmark exam timetabling problems are given in Table I. The statistical results of IICHEA and ICHEA on all the problems from University of Toronto benchmark exam timetabling problems (version I) from [54, 73] are shown in Table II. We only used version I because it has been mostly reported in the literature. ICHEA results are in the brackets. We also compared our best solutions with other published results from [1, 6, 11, 16, 17, 26, 31, 40, 72] sighted frequently in the literature in Table III.

**5.2    Generic fitness function**

Sometimes constraints have hierarchical structure where one constraint is preferred over another. We used all same benchmark timetabling problems with same parameter settings for IICHEA to be solved with a fitness function from Eq. (13). This fitness function is generic and does not have problem dependent weights. The purpose is not to replace weight based penalty function but to show how a generic fitness function can be applied to any constraint problem. The currently form of the equation is for mutually exclusive constraints. Hence we have considered the constraints as only the distance between exams and not the combination of distance and students involved in the exams. We do not have any published results to compare with our results as only problem dependent weights have been used for benchmark exam timetabling problem. We only show the best results from some of the benchmark problems run for ∼1 hour in Table IV.

*TABLE IV Best results of some benchmark exam timetabling problems from IICHEA*

| Instance | $l_0$ | $l_1$ | $l_2$ | $l_3$ | $l_4$ |
|---|---|---|---|---|---|
| Car91 | 1522 | 1604 | 1610 | 1641 | 882 |
| Ear83 | 316 | 372 | 352 | 363 | 200 |
| Hec92 | 124 | 133 | 140 | 140 | 90 |
| Kfu93 | 534 | 511 | 515 | 505 | 232 |
| Lse91 | 409 | 422 | 434 | 464 | 180 |
| Sta83 | 123 | 147 | 123 | 137 | 116 |
| Tre92 | 418 | 453 | 507 | 470 | 272 |
| Ute92 | 184 | 192 | 184 | 250 | 76 |
| Yor83 | 320 | 438 | 352 | 464 | 212 |

## 6 Discussion

Experimental results for benchmark exam timetabling problems for COPs are very promising. Results for problems *Ear83*, *Hec92*, *Sta83*, *Tre92*, *Ute92* are in top three and other results are also in the upper half of the best results. It is noted that IICHEA has been giving consistent results for all the problems. It is noted that exam timetabling problems show good results with hyper-heuristics. Using the incrementality technique of IICHEA on these hyper-heuristics can produce even better results as shown in the comparative results between ICHEA and IICHEA. Incrementality in ICHEA produces better results than without incrementality. Consequently, IICHEA can also be used for real time discrete COPs. IICHEA also does not require having a separate problem specific algorithm to get feasible solutions as a preprocessor for constraint optimization. It has found feasible solutions for all the problems at the SR of 100%. The current version of IICHEA seems to have many parameters, however most of the parameters are logical whose values are selected intuitively that should give similar results to other class of problems as well. We also took the initiative to use generic fitness function for evolutionary search for single objective COPs which can become very handy when confronting a new problem to get initial results without much effort. The current generic fitness functions are quite naïve that caters for mutually exclusive constraints only. The future work is to enhance the functions for other types of constraints as well.

## 7 Conclusion

This paper focuses on incorporating ICHEA for solving discrete COPs. ICHEA has been designed as a generic framework for evolutionary search that extracts and exploits information from constraints. ICHEA has shown promising results experimented on CSPs and COPs. We proposed another version of *intermarriage* crossover operator for discrete CSPs to get the feasible solutions. The experimental result on exam timetabling problems requires additional optimization techniques that are not all generic in its current form. Additionally, it uses many problem specific mutation strategies to optimize it. We also proposed a generic fitness function for single objective COPs that is inspired from developing a *pareto-front* using multi objective COPs. However the status of this fitness function is still in its infancy as it only solves mutually exclusive constraints. A major experimental observation was realizing the efficacy of incrementality in evolutionary search. Incrementality helps in getting feasible solutions with SR of 100% that also produces solutions of better quality. Incremental ICHEA can also be used for real time dynamic COPs in discrete domain. The competitive results from ICHEA shows its potential in making a generic evolutionary computational model that discovers information from constraints. Future work also involves analyzing efficiency and contribution of individual operators towards optimization when several operators are involved in the optimization process. An algorithm with multiple operators like ICHEA generally gives mediocre results when only a single operator is applied in the algorithm, however, collectively with other operator(s) good solutions are obtained. We intend to describe how each operator behaves in search for optimal solution and impact the environment in terms of population diversity, improvement in solutions and genetic drift.

**Appendix**

**1. Time complexity of intermarriage crossover for N-Queen problem.**

Intermarriage crossover takes two parents randomly and produces two progenies. For N-Queen problem first the non-duplicate values in each of the parents has to be found. The corresponding non-duplicate values are appended to each other chromosomes. Number of operations required to find the non-duplicate values in two chromosomes are computed below.

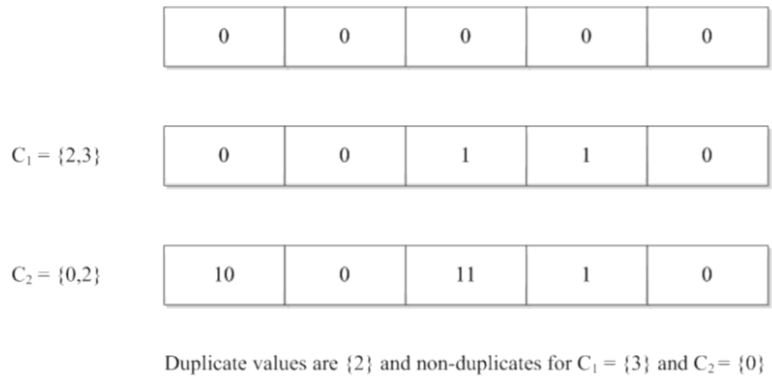

*Fig. A.1 Find non-duplicate values in 2 chromosomes*

A zero vector $A = \langle a_1, a_2, \cdots, a_N \rangle$ of cardinality $N$ is created where $N$ is the size of the N-Queen problem i.e. $A = \mathbf{0}$ and $|A| = N$. Two chromosomes can be represented by $C_1 = \langle c_1, c_2, \cdots, c_{l_1} \rangle$ and $C_2 = \langle c_1, c_2, \cdots, c_{l_2} \rangle$ where $l_1$ and $l_2$ are lengths of chromosomes $C_1$ and $C_2$ respectively.

Set $\forall a_{c_i} = a_{c_i} + 1$ where $\forall c_i \in C_1$ and $\forall a_{c_i} = a_{c_i} + 10$ where $\forall c_i \in C_2$. Fig. A.1 shows an example of finding non-duplicate values for two chromosomes $C_1 = \{2, 3\}$ and $C_2 = \{0, 2\}$. Firstly value 1 is added at location 2 and 3 on vector $A$ for $C_1$ then value 10 is added on the same vector A at location 0 and 2 for $C_2$. This makes $A = \langle 10, 0, 11, 1, 0 \rangle$. Iterating through vector $A$ gives non-duplicate values. Locations of value 1 and 10 in the vector $A$ gives the non-duplicate values for chromosomes $C_1$ and $C_2$ respectively.

The set of non-duplicate values for $C_1$ is: $D_1 = \{x \in A | x = 1\}$ and set of non-duplicate values for $C_2$ is $D_2 = \{x \in A | x = 10\}$. The length of these non-duplicate sets can be given as $l'_1 = |D_1|$ and $l'_2 = |D_2|$. The computational requirement for finding these non-duplicate values can be given as:

$l_1$ iterations are required to add 1 to vector $A$ for the corresponding $c_i$ values of $C_1$ and $l_2$ iterations are required to add 10 to the same vector $A$ for the corresponding $c_i$ values of $C_2$. Then $N$ iterations are required to locate value 11 in the vector $A$. Hence the total operations $T_1$ can be given in Eq. (A.1):

$$T_1 = l_1 + l_2 + N \tag{A.1}$$

The second expression $N(l'_1 + l'_2)$ indicates the operations required to append the non-duplicate allele values to each other parents. Operations required to append $l'_1$ values to $C_2$ and $l'_2$ values to $C_1$ is given in Eq. (A.2) and Eq. (A.3).

$$l_2 + (l_2 + 1) + (l_2 + 2) + \cdots + (l_2 + l'_1) \tag{A.2}$$

$$l_1 + (l_1 + 1) + (l_1 + 2) + \cdots + (l_1 + l'_2) \tag{A.3}$$

Eq. (A.2) and Eq. (A.3) can be resolved as Eq. (A.4) and Eq. (A.5) respectively.

$$l_2 l'_1 + \frac{(l'_1 - 1)l'_1}{2} \tag{A.4}$$

$$l_1 l'_2 + \frac{(l'_2 - 1)l'_2}{2} \tag{A.5}$$

Now total operations $T_2$ required for appending non-duplicate valued to each other is given in Eq. (A.6).

$$T_2 = l_2 l'_1 + \frac{(l'_1 - 1)l'_1}{2} + l_1 l'_2 + \frac{(l'_2 - 1)l'_2}{2}$$

$$\leq l_2 l'_1 + {l'_1}^2 + l_1 l'_2 + {l'_2}^2$$

$$= l'_1(l_2 + l'_1) + l'_2(l_1 + l'_2)$$

$$\leq l'_1(N) + l'_2(N) \tag{A.6}$$

Hence the time complexity $T$ after including both terms from Eq. (A.1) and Eq. (A.6) is given in (A.7):

$$T = T_1 + T_2$$

$$T = (l_1 + l_2 + N) + N(l'_1 + l'_2) \tag{A.7}$$

**2. Preference based penalty function**

When constraints are to be solved according to the preferences then Eq. (11) can be used to solve a given COP. This equation gives higher fitness to the solution where constraints are solved with higher preferences. For example if two solutions $\langle 5, 3, 0, 2\rangle$ and $\langle 5, 4, 0, 1\rangle$ have total of 10 constraints where each element indicates the number of solved constraints with the order of preferences from highest to lowest. For the first solution 5 constraints are solved with the highest preference and 3 constraints are solved with second highest preference and so on. The second solution has better fitness as constraints solved with first order preference is same but the second solution solves more constraints with second order of preference. Eq. (11) can be solved easily with proof by contradiction. Eq. (11) can be restated as:

$$f(l) = \sum_{p=0}^{D} l_p (\mu)^{D-p} \tag{A.8}$$

where $f(l) \Rightarrow$ fitness function for $l$ satisfied constraints

$L \Rightarrow$ Total constraints where $L > 0$

$l_p \Rightarrow$ Total degree of satisfaction with $p^{\text{th}}$ preference.

$\mu \Rightarrow$ A positive constant.

$D \Rightarrow$ Lowest preference defined (highest numeric value for preference) where $D > 0$

$p \Rightarrow$ Preference for constraints in domain $[0\ D]$

To proof by contradiction we use a proposition H1: $f(l) > f(l')$ for $\forall l_k = l'_k$ where $k = \{0,1,2,\ldots,i-1\}$ and $l_i > l'_i$. The idea is to give higher fitness to the solutions that satisfy more constraints with higher preferences. Let's say proposition H1 does not hold for $l_i > l'_i$ i.e $\neg H1 = f(l) \leq f(l')$

$$f(l) \leq f(l')$$

$$\mu^{D-0}(l_0) + \mu^{D-1}(l_1) + \ldots + \mu^{D-i}(l_i) + \ldots + \mu^{D-D}(l_D) \leq$$

$$\mu^{D-0}(l'_0) + \mu^{D-1}(l'_1) + \ldots + \mu^{D-i}(l'_i) + \ldots + \mu^{D-D}(l'_D)$$

Since $l_k = l'_k$ where $k = \{0,1,2,\ldots,i-1\}$ we have:

$$\mu^{D-i}(l_i - l'_i) + \cdots + \mu^{D-D}(l_D - l'_D) \leq 0$$

$$\mu^{D-i}(l_i - l'_i) \leq \mu^{D-D}(l'_D - l_D) + \cdots + \mu^{D-(i+1)}\left(l'_{(i+1)} - l_{(i+1)}\right)$$

So the upper bound of the RHS can be transformed into a geometric series for $\forall l'_k = L^2$ and $l_k = 0$ where $(i+1) < k \leq D$. $L^2$ is the maximum possible value that any $l_k$ or $l'_k$ can have determined through the summation of constraint satisfaction for $k^{th}$ preference $(L-1) + (L-2) + \cdots + 1 = \frac{L(L-1)}{2} < L^2$ of all the constraints. First constraint can make maximum of $(L-1)$ satisfaction with $k^{th}$ preference, second constraint can make maximum of $(L-2)$ satisfaction with $k^{th}$ preference, and so on.

$$\therefore \mu^{D-i}(l_i - l'_i) \leq \mu^0(L^2) + \cdots + \mu^{D-(i+1)}(L^2) = L^2 \frac{(\mu^{D-(i+1)+1} - 1)}{\mu - 1}$$

$$\mu^{D-i}(l_i - l'_i) \leq \mu^{D-i}\left(\frac{L^2}{\mu - 1}\right)$$

$$(l_i - l'_i) \leq \frac{L^2}{\mu - 1}$$

Since the expression $(l_i - l'_i)$ contradicts if it is $< 1$, $\therefore \frac{L^2}{\mu-1}$ should also be $< 1$.

$$\Rightarrow \mu > L^2 + 1$$

We can choose any value for $\mu > L^2 + 1$. One possible option is $L^2 + 2$. Hence:

$$(l_i - l'_i) \leq \frac{L^2}{\mu - 1} < \frac{L^2}{L + 1} < 1$$

Since $l_i > l'_i$ and both are whole numbers so the above statement is contradictory. Let say $l_i = l'_i + x$ where $x \geq 1$ then the above equation can be rewritten as:

$$l'_i + x - l'_i < 1$$

$x < 1$ (Contradiction)

This demonstrates proposition $\neg H1$ does not hold. Hence our hypothesis $H1: f(l) > f(l')$ is true for $\forall l_k = l'_k$

where $k = \{0,1,2,...,i-1\}$ and $l_i > l'_i$. ∎

The Proof of Eq. (12) for minimization of $l_p$ for lower preferences is similar to above. If two solutions $\langle 3, 5, 2, 2 \rangle$ and $\langle 5, 1, 3, 2 \rangle$ have total of 10 constraints where each element indicates the number of solved constraints with the order of preferences from highest to lowest. One constraint can overlap with other constraints with different preferences. The first solution has better fitness as constraints solved with last order of preference is same but the first solution has less constraints of unwanted (high order) preferences. We use a proposition H1: $f(l) > f(l')$ for $\forall l_k = l'_k$ where $k = \{i+1, ..., D\}$ and $l_i < l'_i$. The idea is to give higher fitness to the solutions that minimizes the constraint satisfaction with high order preferences. The generic form of Eq. (12) is:

$$f(l) = \sum_{p=0}^{D}(L^2 - l_p)(\mu)^p \tag{A.9}$$

Let's say proposition H1 does not hold for $l_i < l'_i$ i.e $\neg H1 = f(l) \leq f(l')$

$$f(l) \leq f(l')$$

$$\mu^0(L^2 - l_0) + \mu^1(L^2 - l_1) + ... + \mu^i(L^2 - l_i) + ... + \mu^D(L^2 - l_D) \leq$$
$$\mu^0(L^2 - l'_0) + \mu^D(L^2 - l'_1) + ... + \mu^i(L^2 - l'_i) + ... + \mu^D(L^2 - l'_D)$$

Since $l_k = l'_k$ where $k = \{i+1, ..., D\}$ we have:

$$\mu^0(l'_0 - l_0) + \cdots + \mu^i(l'_i - l_i) \leq 0$$

$$\mu^0(l'_0 - l_0) + \cdots + \mu^{(i-1)}(l'_{(i-1)} - l_{(i-1)}) \leq \mu^i(l_i - l'_i)$$

So the lower bound of the LHS can be similarly transformed into geometric series for $\forall l'_k = 0$ and $l_k = L^2$ where $0 < k \leq i-1$.

$$\mu^0(-L^2) + \cdots + \mu^{(i-1)}(-L^2) \leq \mu^0(l'_0 - l_0) + \cdots + \mu^{(i-1)}(l'_{(i-1)} - l_{(i-1)})$$

$$\therefore \mu^i(l_i - l'_i) \geq (-L^2)\frac{\mu^i - 1}{\mu - 1} = \frac{L^2 - L^2\mu^i}{\mu - 1} \geq \frac{-L^2\mu^i}{\mu - 1}$$

$$\frac{-L^2\mu^i}{\mu - 1} \leq \mu^i(l_i - l'_i)$$

$$(l_i - l'_i) \geq \frac{L^2}{1 - \mu}$$

Since the expression $(l_i - l'_i)$ should be $\leq -1$, it contradicts if it is $> -1$, $\therefore \frac{L^2}{1-\mu}$ should also be $> -1$. Note that $\mu$ is a positive integer that makes the expression $(1 - \mu)$ a negative integer.

$$\Rightarrow \mu > 1 + L^2$$

We can choose any value for $\mu > 1 + L^2$. One possible option is $L^2 + 2$. Hence:

$$(l_i - l'_i) \geq \frac{L^2}{1-\mu} = \frac{L^2}{-1-L^2} = -\left(\frac{L^2}{L^2+1}\right) > -1$$

Since $l_i < l'_i$ and both are whole numbers so the above statement is contradictory. Let say $l_i = l'_i - x$ where $x \geq 1$ then the above equation can be rewritten as:

$$l'_i - x - l'_i = -x > -1 \text{ or } x < 1 \text{ (Contradiction)}$$

This demonstrates proposition $\neg H1$ does not hold. Hence our hypothesis $H1: f(l) > f(l')$ is true for $\forall l_k = l'_k$ where $k = \{i+1, \ldots, D\}$ and $l_i < l'_i$. ∎

The proof of Eq. (13) needs to demonstrate higher functional value for higher value of $l$. The maximum value of the expression $(L^2 - l_p)$ in Eq. (12) is $L^2$ and the maximum possible value for Eq. (12) using sum of geometric series is:

$$f(l) \leq L^2 \frac{(L^2+2)^{D+1}-1}{(L^2+2)-1} < \frac{L^2((L^2+2)^{D+1}-1)}{L^2} < (L^2+2)^{D+1}$$

Using the maximum value derived above Eq. (12) can be modified as:

$$f(l) = (L^2+2)^{D+1} \sum_{p=0}^{D} l_p + \sum_{p=0}^{D} (L^2 - l_p)(L^2+2)^p$$

which can be easily proven by showing proof by contradiction.

### 3. Selecting infeasible solutions

If the population size of infeasible solutions towards the end of a generation is $|pop_{CSP}| + k$ where $k$ is additional infeasible solutions generated through *intermarriage* crossover operator in a given generation then we need to pick the solutions that have good fitness and that also represent the diverse population. Firstly, the population is sorted decreasingly according to the fitness value. The purpose is to retain good solutions with high fitness values to keep the search focus towards better solutions while maintaining diversity by keeping solutions from different regions in the search space even with low fitness values. Mathematically, the selection of solutions is heavy towards the initial indices and lighter towards the higher index of the population set. It can be formulated using the following exponential function in Eq. (A.10):

$$f(i) = e^{-\rho\left(\frac{|POP_{CSP}|-i}{|POP_{CSP}|}\right)} \tag{A.10}$$

Function $f(i)$ gives the desired distribution of the solutions to be selected from the population where $i \in \{1, \dots, |POP_{CSP}|\}$ and $\rho$ defines the level of curve whether extreme, normal or least. Fig. A.2 shows the effect of different values of $\rho$ on the exponential function. Continuous values of function $f(i)$ need to be mapped to discrete indices of the population of sorted solutions which can be done using the following transformation function $f'(i)$ in Eq. (A.11):

$$f'(i) = \left\lfloor (|POP_{CSP}| + k)\frac{f(i)-f(1)}{1-f(1)} \right\rfloor \tag{A.11}$$

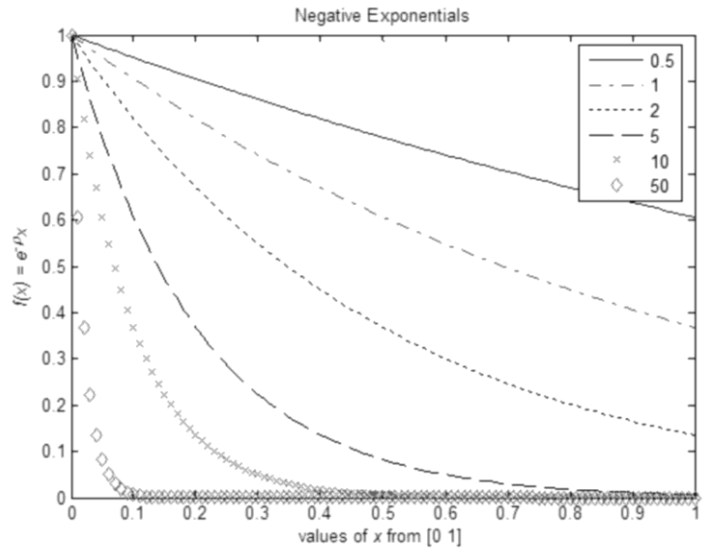

*Fig. A.2 Mapping of population for selection*

The mapping function $f'(i)$ in Eq. (A.11) gives duplicate indices which are again transformed to the nearest indices using function $f''(i)$ in Eq. (A.12) that ensure all unique solutions to be selected from the population.

$$f''(i) = \begin{cases} f'(i-1) + 1, & \text{if } f'(i) \leq f'(i-1) \\ f'(i), & \text{otherwise} \end{cases} \tag{A.12}$$

To see a numerical example using Eq. (A.10) − Eq. (A.12), we choose $\rho = 5$, $|POP_{CSP}| = 10$ and $k = 40$. The result from Eq. (A.10) is:

$$f = \begin{Bmatrix} 0.0067, 0.0117, 0.0205, 0.0357, 0.0622, 0.1084, \\ 0.1889, 0.3292, 0.5738, 1.0000 \end{Bmatrix}$$

These continuous values of function f are transformed into function $f'$ using Eq. (A.11) that gives the indices of the population. It can be noted that the selection of solutions is heavy towards the initial indices of the sorted population.

$$f' = \{0, 0, 0, 1, 2, 5, 8, 15, 27, 49\}$$

The output of function $f'$ has some duplicate values which are not desired so these duplicate values are transformed to the nearest unique values using Eq. (A.12) as shown below:

$$f'' = \{0, 1, 2, 3, 4, 5, 8, 15, 27, 49\}$$

Here the index values start from 0 and ends with $|POP_{CSP}| + k - 1$ which are the first and last elements of the sorted set of infeasible solutions respectively.

## Acknowledgement

This research work is part of the author's PhD work in University of Canberra which is also supported by The University of the South Pacific.